\documentclass[preprint]{elsarticle}
\usepackage{lineno,hyperref}
\usepackage{booktabs}
\usepackage{multirow}
\usepackage{amsmath}
\usepackage{amsfonts}
\usepackage{graphics}
\usepackage{graphicx}
\usepackage{todonotes}
\usepackage{subcaption}
\usepackage{xcolor}
\usepackage{cleveref}

\Crefname{figure}{Fig.}{Figs.}

\journal{Neurocomputing}

\newcommand{\etal}{\textit{et al}. }
\newcommand{\ie}{i.\,e.\ }
\newcommand{\eg}{e.\,g.\ }

\begin{document}

\begin{frontmatter}

  \title{Anomaly Detection in Predictive Maintenance: A New Evaluation Framework for Temporal Unsupervised Anomaly Detection Algorithms}
  \tnotetext[mytitlenote]{}

  \author[dasci]{Jacinto Carrasco}
  \ead{jacintocc@decsai.ugr.es}

  \author[dasci]{David López}
  \author[dasci]{Ignacio Aguilera-Martos}
  \author[dasci]{Diego García-Gil}

  \author[arcelor]{Irina Markova}
  \author[arcelor]{Marta García-Barzana}
  \author[arcelor]{Manuel Arias-Rodil}

  \author[dasci]{Julián Luengo}
  \author[dasci]{Francisco Herrera}

  \address[dasci]{Department of Computer Science and AI, Andalusian Research
    Institute in Data Science and Computational Intelligence, University of
    Granada, Granada, Spain }
  \address[arcelor]{ArcelorMittal Global R\&D, New Frontier, Digital Portfolio}

  \begin{abstract}
    The research in anomaly detection lacks a unified definition of what
    represents an anomalous instance. Discrepancies in the nature itself of an
    anomaly lead to multiple paradigms of algorithms design and experimentation.
    Predictive maintenance is a special case, where the anomaly represents a
    failure that must be prevented. Related time series research as outlier and
    novelty detection or time series classification does not apply to the
    concept of an anomaly in this field, because they are not single points
    which have not been seen previously and may not be precisely annotated.
    Moreover, due to the lack of annotated anomalous data, many benchmarks are
    adapted from supervised scenarios.

    To address these issues, we generalise the concept of positive and negative
    instances to intervals to be able to evaluate unsupervised anomaly detection
    algorithms. We also preserve the imbalance scheme for evaluation through the
    proposal of the Preceding Window ROC, a generalisation for the calculation
    of ROC curves for time series scenarios. We also adapt the mechanism from a
    established time series anomaly detection benchmark to the proposed
    generalisations to reward early detection. Therefore, the proposal
    represents a flexible evaluation framework for the different scenarios. To
    show the usefulness of this definition, we include a case study of Big Data
    algorithms with a real-world time series problem provided by the company
    ArcelorMittal, and compare the proposal with an evaluation method.
  \end{abstract}


\end{frontmatter}

\section{Introduction}
\label{sec:introduction}

The label of anomaly detection is assigned to a variety of problems with
different natures and use cases~\cite{2009-Chandola-Anomalydetectionsurvey}. For
instance, time series anomalies in predictive maintenance or fault detection
\cite{2019-Wang-Outlierdetectionbased}, process data in security
\cite{2020-Zohrevand-DynamicAttackScoring} or graph data in social media
\cite{2018-Noorossana-overviewdynamicanomaly}. Other names like outliers,
exceptions, rare events or novelties are used with different intention or in
different study fields. The disparity of scenarios itself develops into multiple
evaluation schemas, that may cause confusion and proposals that are not
correctly evaluated.

The variety of nomenclature to the scenarios has been addressed by other
researchers~\cite{2019-Carreno-Analyzingrareevent}. Their proposed taxonomy uses
\textit{rare event} for supervised temporal data with a class imbalance and the
task of classifying these time series into known classes
\cite{2016-Theofilatos-PredictingRoadAccidents}. The broadly used term of
anomaly is reserved for the supervised classification task of non-temporal data
with highly imbalanced class distribution
\cite{2016-Ribeiro-Sequentialanomaliesstudy}. In the semi-supervised
scenarios~\cite{2017-Aggarwal-OutlierAnalysis}, where only normal data is
available during training time, some authors use the term of \textit{One-Class}
classification~\cite{2009-Chandola-Anomalydetectionsurvey},
\cite{2020-Fernandez-Francos-OneClassConvexHullBased}, while others prefer
\textit{Novelty detection}, remarking the interest in the unseen instances
\cite{2008-Oliveira-Noveltydetectionconstructive}. Another broadly used term is
\textit{outlier detection}~\cite{2017-Aggarwal-OutlierAnalysis}, usually associated with
unsupervised classification and frequently in relation with the term
\textit{noise}, more related to the data instances that divert from normal
observations but not enough to be considered to have been produced by another
mechanism, which is in the ultimate instance the aim of abnormal data detection
algorithms.

In the specific scenario of predictive maintenance, as a special case of anomaly
detection in time series and the situation considered in this work, the events
of interest are represented by singular points that require an
intervention. A similar circumstance happens with time series anomaly detection
algorithms: they evaluate singular points to predict an event that does not
affect only an instant but a subsequent interval.

As expected, all the mentioned different scenarios use
specific measures and experimentation setups to validate their results. On the
one hand, most of these measures come from supervised
imbalanced scenarios, as the recall of the minority or new classes, or some
combination of the precision and recall, like the F-measure, or measures derived
from the ROC (\textit{Receiver Operating Characteristic}) Curve, such as the
Area under the curve
(AUC)~\cite{2016-Goldstein-ComparativeEvaluationUnsupervised}. On the other
hand, for those scenarios where the main focus is the temporal component, the
proposals include a measure for the earliness of the
detection~\cite{2015-Lavin-EvaluatingRealTimeAnomaly} with the
  aim of minimising the probability of a failure.

In summary, there are many different scenarios for abnormal behaviour that go by
the name of anomaly detection and many associated measures which have associated
drawbacks with respect to the addressed task of predictive
  maintenance:

\begin{itemize}
  \item The benefits of the imbalance problem approach are diluted by the
        omission of the temporal component.
  \item The evaluation of the earliness in the detection is aggregated for a
        more understandable comprehension through more parameters, making more
        difficult the task of adaptation to general and real case scenarios.
\end{itemize}

%

To overcome these issues, we gather the different approaches to anomalous event
detection, unifying the evaluation schemes and the implications of the instance
and class assignation into an new evaluation framework for temporal unsupervised
anomaly detection algorithms for scenarios where the events of interest have a
uncertain relation with the data. The proposed evaluation framework definition
starts with a temporal window that precedes the anomaly. The
general consideration is that a positive instance is detected when the algorithm
triggers an alarm within the window. However, this raises some logic issues:
What is the upper window limit? What is the usefulness of windows so wide that
they gather all the instances? Theoretically, we can consider the whole interval
as the previous window. Then, we would have only a positive instance relative to
the last logged event, and negative instances posterior to this last instance,
which lack a relevant meaning for a study, as random detectors would benefit
from these schemes. Nonetheless, the upper limit is subject to the researcher
interests. The proposed framework could be outlined using the following
components:

\begin{itemize}
  \item In the first place, we define a transformation of the instances, from
        time-stamps to intervals, with an aggregation that provides a wider view
        of the event of an anomaly through the inclusion of a parameter for the
        length of the window that precedes each annotated event.
  \item Then, we describe what are the options for the aggregation and what are
        the implications, including an earliness-aware aggregation derived from
        the Numenta Anomaly Benchmark (NAB).
  \item The next component is the Preceding Window ROC (pw-ROC), that generalise
        the definition of the classic ROC curve to include the previously
        mentioned window length parameter. The aggregation of these pw-ROC
        shapes the ROC surface, as the parameter of the length of the preceding
        window represents the third dimension.
\end{itemize}

Therefore, the complete process generates a figure that provides the information
about the quality of the algorithm, not only at all the
  possible levels of threshold to assign the anomaly labels but also to
different levels of distance until the event of interest. Two versions of this
proposal has been implemented: one classical version using python, and another
version using pySpark to be capable of handling Big Data time series
problems. 
This proposal has two implementations in python, one classic implementation and
a distributed version using pySpark to be capable of handling Big Data
time series problems and is available as a repository in
GitHub\footnote{\url{https://github.com/ari-dasci/S-pwROC}}.

To validate the usefulness of the proposal, we include the evaluation using this
framework of three state-of-the-art algorithms and examine their results using a
real data set provided by
ArcelorMittal\footnote{\url{https://corporate.arcelormittal.com/}}. We also
evaluate the algorithms using a scoring system for anomalous range and compare
this evaluation with our method to analyse the benefits of the proposal. The
popularity of the scenario, where the use of unsupervised algorithms is desired
due to the lack of certainty about the possible events that arise, although the
events of interest can be annotated through observation for algorithms
evaluation, shows the reliability of the proposed benchmark for such task.

In summary, the major contributions of this work are:
  \begin{itemize}
    \item Description of the distinctive features of the anomaly detection problem for time series scenarios of predictive maintenance.
    \item Proposal of evaluation method for the described scenarios with associated software for Big Data time series.
    \item Case study with a comparison between the outcome of the evaluation method proposal and an evaluation proposal for a similar scenario.
  \end{itemize}

This paper is organised as follows:~\autoref{sec:backgr-eval} presents the
current state of anomaly detection evaluation systems, approaches and quality
measures. In \autoref{sec:methodology}, the proposed evaluation framework is
presented and justified theoretically. \autoref{sec:results} includes the
experiments and comparison performed to validate the applicability and validity
of the proposal. Finally, \autoref{sec:conclusion} concludes the paper.

\section{Background in anomaly detection evaluation}
\label{sec:backgr-eval}

In this section, we describe four different approaches for the evaluation of
anomaly detection algorithms, particularly for those situations where temporal
component should be taken into account. The most used strategy is to consider
the problem of anomaly detection (without a temporal component) as an imbalanced
classification task. In predictive maintenance the events of interest can be
transformed not into positive instance in a classification problem but into the
target variable for a regression algorithms perspective. More recent works
shift back to classification tasks with extra measures to reward the desired
earliness in the detection. For all these methodologies, the underlying anomaly
detection aim is to discern the timestamp when an anomaly occurs using the
observations $x_i, i=1,\dots, N$, \ie to provide an accurate label $y_i$ for
each instance that reflects the time series behaviour.

\subsection{Anomaly detection evaluation for non-temporal data}
\label{sec:non-temp-eval}

For methods that provide a ranking of the instances according to their outlierness, the precision at $n$ ($P @ n$) is defined as the proportion of the first ranked $n$ instances that are anomalies~\cite{2009-Craswell-Precision}. If $n$ is equal to the number of outliers in the dataset, the the author denominates $P @ n$ as the $R$-precision. This makes the reliability of this measure compromised by the $n$ parameter, especially in unsupervised scenarios \cite{2019-Xu-RecentProgressAnomaly}. The $P @ n$ measure would ignore the temporal component in predictive maintenance scenarios, considering the observations as isolated instances.

The problem of the balance is addressed in imbalance classification with the ROC
curve \cite{2018-Fernandez-LearningImbalancedData}. ROC space is defined as an
$[0,1]\times[0,1]$ space using True Positive Rate (TPR or sensibility, represented in
$Y$ axis) and False Positive Rate (FPR or $1-$ specificity, represented in $X$
axis) \cite{1982-Hanley-meaningusearea}. For example, an algorithm with a
perfect TPR and FPR would be in $(0,1)$. The extension of this concept is used
generally for algorithms that provide scores or probabilities in imbalanced
classification scenarios. Then, we can get different ROC points for each
possible threshold and TPR (and FPR) increases as this threshold does.

\subsection{Regression transformation}
\label{sec:regression}

As described in \autoref{sec:introduction}, temporal data represents a great
proportion of real-world problems. The most common task in time
  series problems is to model the desired time series, a scenario where the
  temporal component is intrinsic to the studied data, so the evaluation measure
  does not need to be aware of this temporal nature and they use a regression
  measure like the Root Mean Squared Error (RMSE) or the Error Ratio
  \cite{2013-Zeng-Timeseriesmodeling, 2016-Zeng-Inferringnonlinearlateral}.
  However, for the specific case of anomaly detection in time series, the
  interest relies on the anomalous events that take place at a certain point in
  the time series, not the values of the time series themselves.

One approach for the evaluation of anomaly detection in time
  series comes from the transformation of the problem of rare event detection
into a regression problem, where for each timestamp $i = 1, \dots, N$, the
target is the remaining time $r_i$ until the failure or the
stop~\cite{2019-Gunay-inquirypredictabilityfailure}. This is an interesting
proposal, as in predictive maintenance we aim to maximise productivity with
minimal repairs costs through the knowledge of our system. Then, reliable
predictions of the available time until the failure
($R(x_i)=\hat{r}_i, i=1,\dots,N$) may help us in this task. The evaluation
measure is the RMSE, as in many regression problems. This approach broaches the
relevance of time in failure detection, and using the RMSE as the quality
measure overcomes the detachment between the singular predictions and the
continuous nature of the problem.

\subsection{Rare event detection and earliness}
\label{sec:rare-event-detection}

In the time series classification problem scenario, the earliness is defined as
the mean percentage of the length $t_j$ of each time series $X_j$ needed to
provide a class label $Y_j$. The pertinence of taking into account the distance
until the predicted event is also addressed by Zhang
\etal~\cite{2017-Zhang-Deeplearningsymbolic}. In this work, they transform the
rare event detection problem into a classification one through the definition of
a horizon window, in which the instances must be classified as anomalous. A
weight is assigned to these positive instances to give more relevance to these
closer to an event.

\subsection{Range-Based Anomaly Detection Measures}
\label{sec:range-based-anomaly}

A relevant scenario for temporal anomaly detection is the consideration of time
intervals as anomalies \cite{2019-Tatbul-PrecisionRecallTime,
  2018-Lee-PrecisionRecallRangeBased}. In these works, Tatbul \etal
propose a range-based generalisation for the concepts of precision, recall and
the $F_1$ measure. The expected output of the algorithm is a label for each
instance, that is transformed to intervals of contiguous anomalous instance.
These measures evaluate each anomalous interval ($R_i$ for real anomalous
intervals and $P_i$ for predicted anomalous intervals) using the score based on
the detection ($E(R_i, P)$, the existence of at least a timestamp that
belongs to $R_i$ and any $P_i \in P$), and the overlap ($O(R_i, P)$, the
proportion of detected timestamps in $R_i$). Therefore, the subject of interest
of this method are anomalous intervals, labeled as such. Their proposal allows
the modification of certain parameters to reward different behaviours, such as
an early or late detection of the anomaly within the real anomalous interval.

This measure has a straightforward modification to evaluate algorithms for the
studied scenario, defining the anomalous time intervals as the windows preceding
the event of interest. Then, to validate our proposal we compare it with this
evaluation method.

\subsection{Numenta Anomaly Detection Benchmark}
\label{sec:numenta-anom-detect}

A vital problem for time series anomaly researches is that there are no extended
benchmarks for the comparison of the performance of streaming anomaly detection
algorithms. A fundamental proposal is the Numenta Anomaly
Benchmark~\cite{2015-Lavin-EvaluatingRealTimeAnomaly}. This benchmark proposes a
scoring function and provides a set of manually labelled real-world time series.
Here, four weights $A_{TP}, A_{FP}, A_{TN}, A_{FN}$ are defined for true
positives, false positives, etc. The absolute value of these weights are between
0 and 1, being $A_{TN}$ and $A_{TP}$ positives and $A_{FN}, A_{FP}$ negatives,
to penalise the errors. The default scenario described in the Numenta Anomaly
Benchmark defines the weights $A_{TP}=A_{TN}=1$, and $A_{FP}=A_{FN}=-1$,
although different values could be settled for different profiles. Then, a
window is defined previous to each anomaly. In these windows, only the first
alarm provided by the algorithm is kept, while the windows
  that precedes an anomaly that do not have an alarm are considered to be false
  negative instances, that could be denoted as missed windows. Let $y$ be the
relative position of the alarm within the interval. Then, the score is defined
as:
\begin{equation} \sigma^A(y) = (A_{TP} - A_{FP}) \left( \frac{1}{1+e^{5y}} \right) - 1. \end{equation}
Therefore, the detections at the end of the interval ($y=0$) are evaluated as 0,
and a detection just after the interval receives a smaller penalisation. Then,
the raw score for each data set is defined as the sum of the scores of the
detections in the positive windows ($Y_{d}$) plus the product of $A_{FN}$ and
the number of missed windows, $f_d$:
\begin{equation} \mathrm{Score}: \sum_{y \in Y_d} \sigma^A(y) + A_{FN}f_d. \end{equation}

This benchmark is subject to some criticisms made by Singh and
Olinsky~\cite{2017-Singh-DemystifyingNumentaanomaly}. Some of these criticisms
refer to the impracticality of the system in a real-world streaming scenario as
the allegedly good performance of the studied algorithms is not enough for
practical applications. Moreover, there are some issues concerning the scoring
function, as it is not clearly defined for every situation. In their analysis,
it is shown that their score over-reward avoiding false positives and allows a
low recall of the anomalies. We address these issues concerning the scoring
system through the combination of an imbalanced scenario metric as the AUC with
the window partition of the temporal space and the weighting system for
rewarding early detection.

\subsection{Summary of quality measures}
\label{sec:summ-qual-meas}

A summary of the different scenarios and used quality measures is included in
\autoref{tab:evaluation}.

\begin{table}[h]
  \centering
  \begin{tabular}{l|p{5cm}|l|l}
    \toprule
    Task & Description & Evaluation & Formula \\
    \midrule
    \multirow{5}{*}{time series class.} & Classification of complete time series. & Accuracy & $\frac{\sum_{X \in \mathbf{X}_{test}}\mathbb{I}[c(X)=Y]}{|\mathbf{X}_{test}|}$ \\
         & Percentage of the length of the time series needed for the classification. & \multirow{3}{*}{Earliness} & \multirow{3}{*}{
                                                                                                                      $\frac{\sum_{x \in \mathbf{X}_{test}}\frac{t^{*}_x}{L}}{|\mathbf{X}_{test}|}$}\\
    \hline
    Regression & Numeric time until stop. & RMSE & $\sqrt{\sum_{i=0}^N(\hat{r}_i-r_i)^2}$\\
    \hline
    \multirow{2}{*}{Outlier/Rare-event} & \multirow{2}{*}{Unsupervised/Supervised} & AUC & Area under ROC curve \\
      & & $P@n$ & $\frac{|\{\sum_{i=1}^n\mathbb{I}[y_{i}=O]\}|}{n}$\\
    \hline
    \multirow{2}{*}{Range-based measures} & \multirow{2}{*}{Supervised Ranges} & Precision/ & $\alpha E(R_i, P) + $ \\
      & & Recall &  $\quad(1-\alpha) O(R_i, P)$\\
    \hline
    \multirow{3}{*}{NAB} & \multirow{3}{=}{Custom scoring system. Detection within weights in windows previous to anomalies.} &
                           \multirow{3}{*}{Custom score} & $\sigma^A(y) = \frac{A_{TP} - A_{FP}}{1+e^{5y}} - 1$ \\
         & & & \\
         & & & $\sum_{y \in Y_d} \sigma^A(y) + A_{FN}f_d$ \\
    \bottomrule
  \end{tabular}
  \caption{Evaluation measures for abnormal behaviour detection}
  \label{tab:evaluation}
\end{table}

\section{Evaluation framework for temporal unsupervised anomaly detection}
\label{sec:methodology}

This section is devoted to the description of the proposal of an evaluation
framework for the time series anomaly detection scenarios found in the
literature so the researchers can obtain more relevant measures according to
their data. As we have mentioned in the previous section, often time series
anomalies are labelled as singular points and scoring systems rewards an early
detection. However, these systems lack a balance mechanism that takes into
account an appropriate weight of the proportion of positive instances. Hence,
they overestimate the relevance of both premature and correct detections to the
detriment of a big amount of false positives or false negatives instances.

We aim to introduce the mechanism of the ROC curve, that takes into account the
relation between the TPR and FPR for the possible thresholds. An additional
benefit of using the ROC curve is the possibility of working with anomaly scores
instead of labels for the predicted instances. This feature provides more
options when designing the combination methods.


This section is structured as follows: In \autoref{sec:first-step:-transf} we
include the formal definition of the proposed transformation of time-stamp
instances into intervals. The aggregation component is described in
\autoref{sec:second-step:-aggr}. \autoref{sec:third-step:-roc} is dedicated to
the definition of the proposed Preceding Window ROC and the considerations of
the resulting ROC surface.

\subsection{First component: Transformation into interval instances}
\label{sec:first-step:-transf}

Let $\mathbf{X} = \{\mathbf{x}_1, \dots, \mathbf{x}_N\}$ be a time series and
$\{t_1, \dots, t_N\}$ be the time-stamps for those instances. Let
$\mathbf{S}=\{s_1, \dots, s_M\}$ be the set of time-stamps where occur the $M$
events. According to the desired study and the scenario, the $M$ incidents could
be time-stamps that represents the events or the start of time intervals. Let
$W_{max} = \min_{i=1,\dots,M-1} \{s_{i+1}-s_{i}\}$ be the maximum length of the
window that defines the positive intervals.

From now on, we denote $w$ as the selected length of the window used to
determine the evaluation intervals and the quality measures. This parameter
belongs to $\left(0,W_{max}\right]$. We could let $w$ be 0, although a
classic time-stamp evaluation would be preferable for such length value.

Concerning the practical value of $w$, we could define the start of the positive
interval for a supervised scenario, as a rule of thumb, as 10\% of the studied
period divided by the number of
anomalies~\cite{2015-Lavin-EvaluatingRealTimeAnomaly}. This value is subjected
to the researchers' and the domain experts' interests, especially for
unsupervised scenarios.

With the previous definitions, we define the set formed of the positive and negative intervals that represent the aggregated instances, which contain the original instances $\mathbf{x}_{j}, j=1,\dots, N$:

\begin{equation} \mathcal{I}_{w} = \{ \{\mathbf{x}_j : s_{i-1} < t_j; s_i-(k+1)w < t_j \leq s_i - kw \} : i = 1, \dots, M; k \in \mathbb{N}_{0} \}, \end{equation}
with $s_{0} = \min\{s_1, t_1\},$ undefined otherwise for $i=1$, and
$\mathbb{N}_{0}$ represents the natural numbers and the 0.


From these instances, we can define in a simpler way the positive instances as
a subset of $\mathcal{I}$, which are the instances within a
distance lower than $w$ from an event:

\begin{equation} \mathcal{P}_{w} = \{ \{\mathbf{x}_{j} : 0 \leq s_{i} - t_{j} < w \} : i = 1, \dots, M \}, \end{equation}
and the negative instances as those with a distance greater than $w$ (from the
right) to an incident:

\begin{equation} \mathcal{N}_{w} = \{ \{\mathbf{x}_{j} : s_{i-1} < t_{j}; s_{i}-(k+1)w < t_{j} \leq s_{i}-kw \} : i = 1, \dots, M; k \in \mathbb{N} \}. \end{equation}

\begin{figure}[h]
  \centering
  \includegraphics[width=\textwidth]{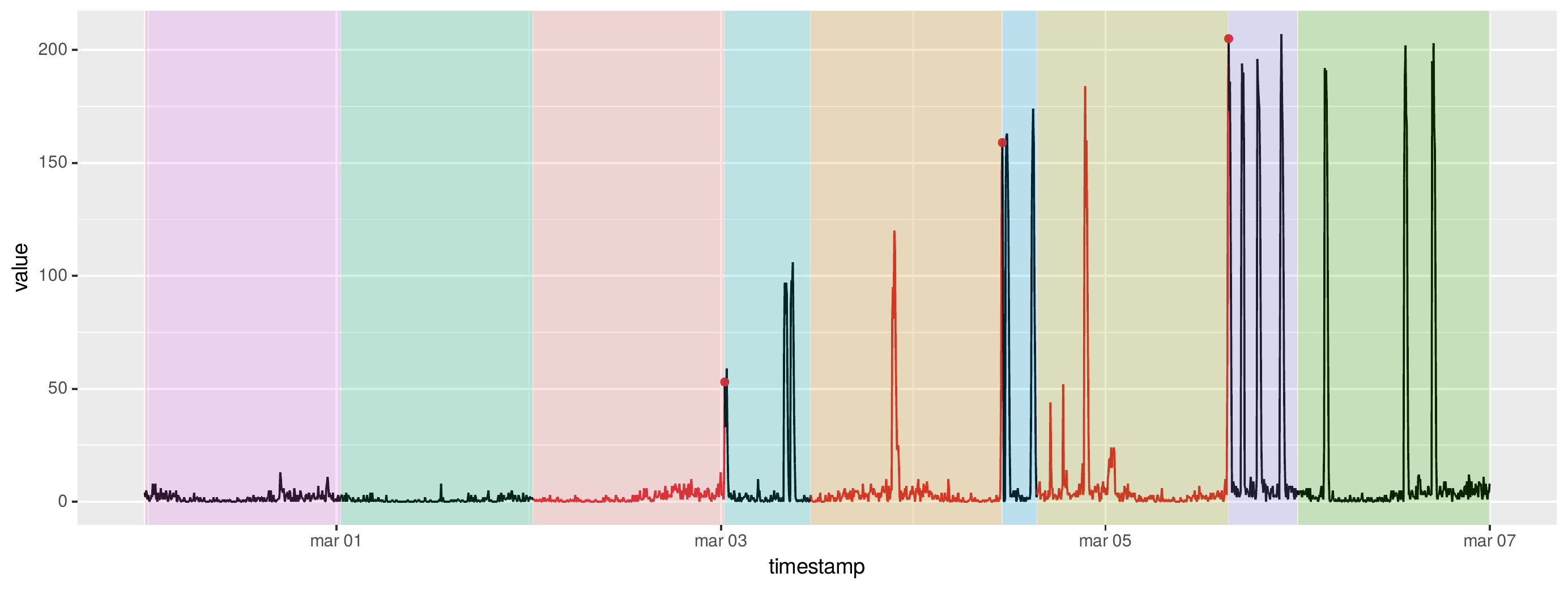}
  \caption{Interval partition of the time series based on the red-dot highlighted
    anomalies.}
  \label{fig:ts-partition}
\end{figure}

From these definitions, it is clear that
  $\mathcal{I}_{w} = \mathcal{P}_{w} \cup \mathcal{N}_{w}, \mathcal{P}_{w} \cap \mathcal{N}_{w} = \emptyset$.
For simplicity, we can denote $\mathcal{I}_w = \{X_l, l=1,\dots, p\}$, where
each $X_l$ represents an aggregation of the original
instances. It is important to note that this definition allows the framework
application to non-uniformly sampled time series data. In
\autoref{fig:ts-partition} it is shown the partition
$\mathcal{I}_{w}$ of a time series in the considered instances
using the color in the background, which would represent the
elements of $\mathcal{I}_w$. The anomalous windows
instances of $\mathcal{P}_w$) are marked with a red line and
the annotated events are marked with a red dot.

\subsection{Second component: Aggregation functions and earliness-aware scoring}\label{sec:second-step:-aggr}



Once the time-stamps are aggregated into interval instances, we propose the use
of a real valuated function $f$ to summarise the anomaly score
provided by the algorithm $A$ for those intervals:

\begin{equation}
  f(X_l) = f\left(\left\{ A(\mathbf{x}): \mathbf{x} \in X_l\right\}\right).
\end{equation}
There are some suitable options for $f$, that depend on the research interest
and the used algorithm. Here we include the main choices:

\begin{description}
  \item[Average] For those algorithms that provide an anomaly score, the mean is
        the basic aggregator. The results are expected to be more representative
        for the scenarios with more instances in each interval.
  \item[CCDF] The Complementary Cumulative Density Function, with a threshold,
        to compute the percentage of instances with an anomaly score greater
        than such threshold. This aggregation function can be suitable for
        scenarios with less instances within each interval or for those
        algorithms that only provide a label instead of an anomaly score. This
        aggregator with a 0.5 threshold is the median aggregation function.
  \item[NAB] The Numenta weighting scheme is reformulated as another aggregation function, giving less relevance to those time-stamp instances that are too close to the event, as it is imminent and an anomaly prediction lacks use.
        \[ f_{NAB, w}(X_l) = \sum_{\mathbf{x}_j \in X_l}\sigma_w(\operatorname{time\_until\_next\_alarm}(t_j)) \cdot \operatorname{algorithm}(\mathbf{x_j}),\]
        where $\sigma_w(t) = \frac{2}{1+e^{15t/w}} - 1$. This gives a weight of near 1 for those instances further to the next event than $w$, so we can ignore the weight in the negative intervals. The 15 coefficient for the distance is derived from Numenta Anomaly Benchmark, as they use a 5 coefficient for a 3 hours window.

\end{description}

\paragraph{Filtering consideration}\label{sec:filt-cons}

For those algorithms that provide an anomaly label for each instance, particularly
for those situations where they provide a high rate of positive labels, the
aforementioned aggregation functions may lead to a high false positive ratio.
As described in the previous section, there are some options for filtering the
positive instances:

\begin{description}
  \item[Non-trigger window] A second window $w_2$ may be defined after each
        instance declared to be positive by the algorithm, within which no
        other positive instance is considered to the aggregation
        \cite{2020-Iturria-otsadpackageonline}.
  \item[Counter] We could considered a sliding window, where only the last
        triggered alarm is declared as an anomaly if there were more than a
        certain number $K$ of alarms in such period.
\end{description}

The use of these functions is independent of the aggregation function and may be
considered as a part of it via composition $f' = f \circ g$, where $g$ represent a
filtering function and $f$ and $f'$ are aggregation functions.

\subsection{Third component: Evaluation based in the Preceding Window
  ROC}\label{sec:third-step:-roc}

ROC curve can be defined as a plot of the sensitivity versus
$(1-\operatorname{specificity})$ for all possible threshold values $c$
\begin{equation} \operatorname{ROC} = \left\{ (P(y > c | \hat{y}=0), P(y > c | \hat{y}=1)): c \in \left(\infty, \infty\right) \right\}, \end{equation}
where $y$ represents the score and $\hat{y}$ the real class. In our proposal,
this definition is subject to $w$, as it determines the score $y$ through the
aggregation made by $f$. The inclusion of the time as a parameter for the
definition of ROC was also suggested by Heagerty \etal
\cite{2000-Heagerty-TimeDependentROCCurves} for a different scenarios. Their
work proposes an estimation for the ROC in a time-stamp posterior to a medical
treatment, and the class represents the survival of the patient. This means that
only the ROC computation is changed with the window length parameter, and it
does not affect the label itself. Moreover, the window of study is posterior to
the interest event and the class of the instance will not ever change again,
unlike in our work, where the windows precedes the event and the class is
negative after the event. For our evaluation framework, we propose the following
definition of the pw-ROC:

\paragraph{Definition} Preceding Window ROC for the window length $w$:
\begin{equation}
  \operatorname{pw-ROC}_w =
  \left\{ (P(f(X) > c | X \in \mathcal{N}_w), P(f(X) > c | X \in \mathcal{I}_w)):
    c \in \left(-\infty, \infty\right) \right\}
\end{equation}

It is important to note the influence of $w$, as it affects the
  definition of $\mathcal{N}_{w}$ and $\mathcal{I}_{w}$. Once we have obtained
  the different pw-ROC curves for the desired window lengths, we can generate a
  ROC surface to observe the difference in the balance between precision and
  specificity concerning the window length. The study of this parameter allows
  the characterisation of the problem if the researchers do not have prior
  information about the possible length of the window where the anomalies can be
  detected in the sensors data. The AUC is expected to improve as the window
length increases as there are fewer negative instances. Therefore, the window
length when this performance surge happens is another element to take into
account when comparing algorithmsIf there are algorithms that
  have a significative better AUC for shorter windows it implies that the
  anomalous events are detectable by these algorithms. Another option is that
  the surge happens in the upper limit of the range of window lengths, which may
  be caused if there are no negative instances, so the expert knowledge may help
  to avoid this issue.

\begin{figure}[h]
  \centering
  \includegraphics[width=0.9\textwidth,clip]{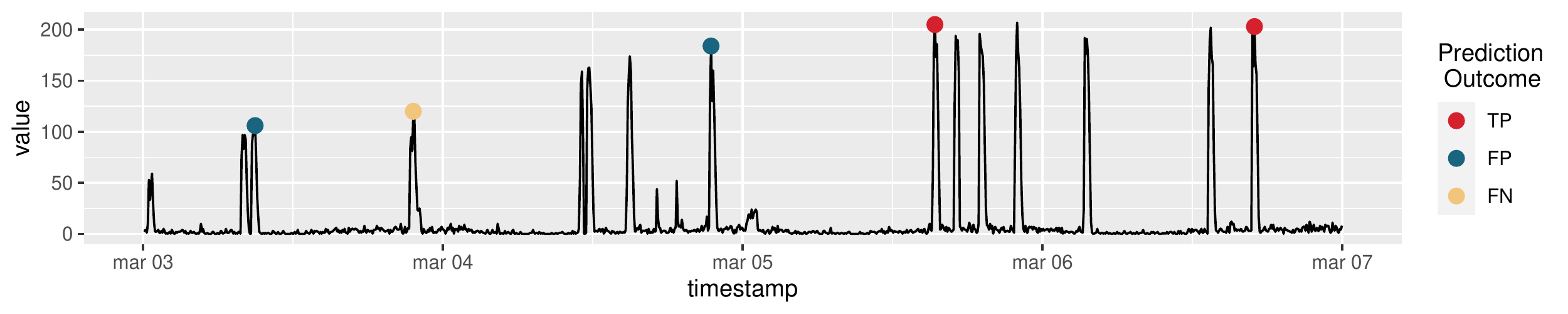}
  \caption{Anomaly scores and algorithm predictions}\label{fig:outcome}
\end{figure}

\begin{figure}[h]
  \centering
  \begin{subfigure}{0.9\textwidth}
    \includegraphics[width=\textwidth,clip]{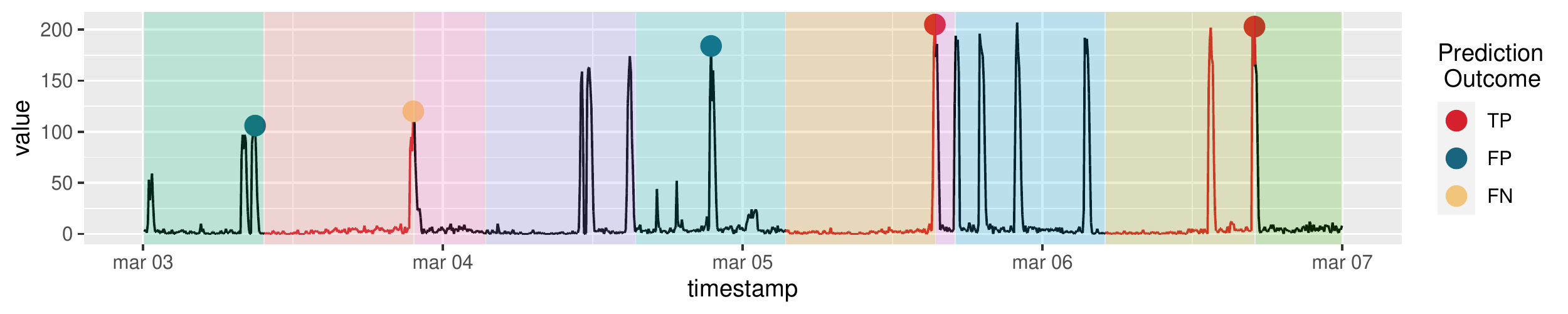}
    \caption{12-hours window}\label{fig:time-partition-2}
  \end{subfigure}
  \begin{subfigure}{0.9\textwidth}
    \includegraphics[width=\textwidth,clip]{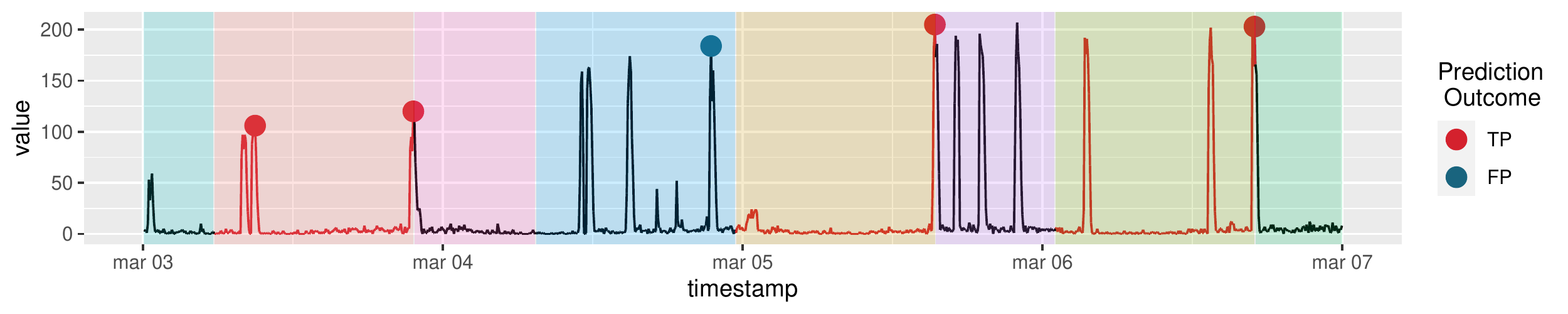}
    \caption{16-hours window}\label{fig:time-partition-3}
  \end{subfigure}
  \begin{subfigure}{0.9\textwidth}
    \includegraphics[width=\textwidth,clip]{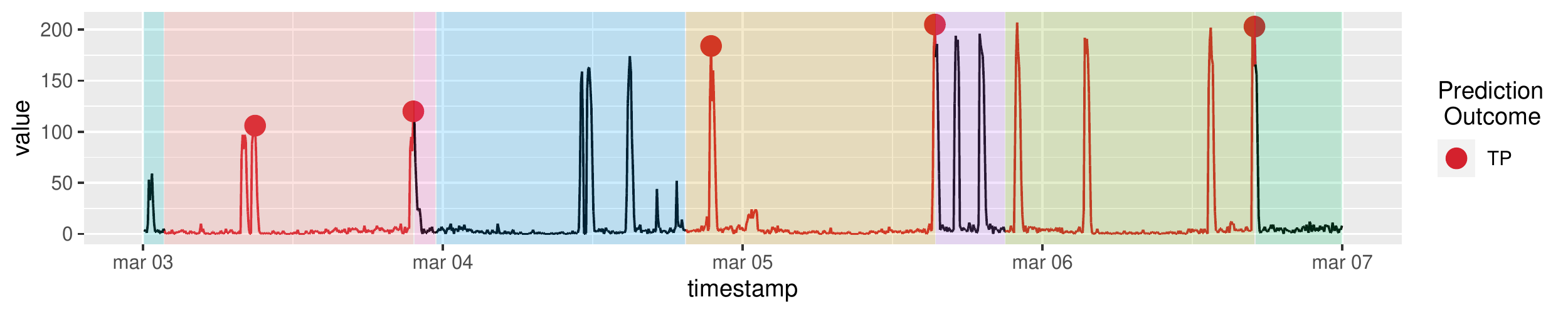}
    \caption{20-hours window}\label{fig:time-partition-4}
  \end{subfigure}
  \caption{Window partitioning}\label{fig:time-partition}
\end{figure}

\begin{figure}[h]
  \centering
  \includegraphics[width=0.7\textwidth,clip]{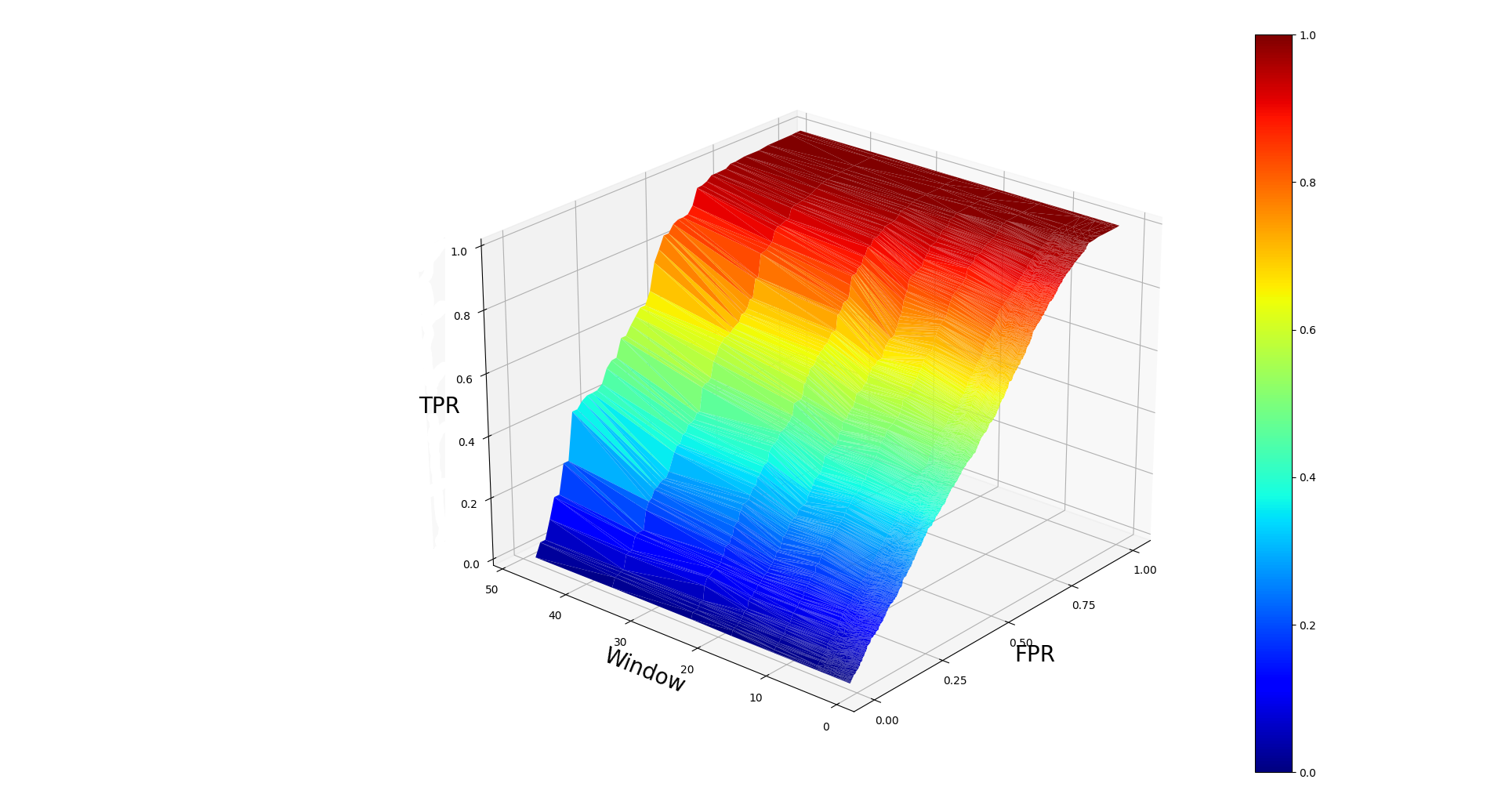}
  \caption{ROC Surface}\label{fig:roc-surface-summary}
\end{figure}

The whole process is summarised in
  \Crefrange{fig:outcome}{fig:roc-surface-summary}. From the anomaly scores or
  labels provided by the algorithm, depicted in \autoref{fig:outcome}, we can
  filter and replicate the window partitioning according to different values of
  the window parameter, obtaining multiple time series partitions, illustrated
  by \autoref{fig:time-partition}. Then, an aggregation is performed depending
  on the researcher interest and a ROC surface is obtained
  (\autoref{fig:roc-surface-summary}).

\autoref{fig:time-partition} also helps to explain the
  influence of the window length in the definition of the anomaly. For the sake
  of simplicity, let consider that the detection algorithm provides only
  $\{0, 1\}$ labels and that the aggregation method is the maximum function.
  Then, an interval is labelled as positive if there is an anomaly predicted in
  such interval. As previously stated, the score is expected to increase with
  longer window lengths, because those undetected anomalous intervals (plotted
  in red and ended with a False Negative dot) could include some positive
  instances that were False Positives with a shorter window, as seen in the
  first two highlighted points in \autoref{fig:time-partition-2} and
  \autoref{fig:time-partition-3}. Similarly, most false positives instances will
  be captured by true positive intervals, increasing the precision, as
  illustrated by the changes between \autoref{fig:time-partition-3} and
  \autoref{fig:time-partition-4}.

\subsection{Software}
\label{sec:software}

The package \texttt{pwROC} implements the described algorithm evaluation method.
As previously stated, this package includes two versions of the method: a non
distributed version, which can be installed without the Spark and pySpark
dependencies, and a distributed version (the sub-module \texttt{pwROCBD}) for Big
Data time series, which requires a working spark installation and has pySpark as
python package dependency.

The package contains the functionality to preprocess the data set, filtering the
instances according to the maintenances, computing a specific ROC curve for a
window length or computing the ROC surface for the desired window lengths. The
available aggregation functions are the mean, median, complementary cdf and the
NAB weighting schema. The \texttt{pwROCBD} sub-module, which has the same functionality as
the classic implementation, requires

By using \texttt{pandas.DataFrames} and \texttt{pyspark.DataFrames},
\texttt{pwROC} enables users to integrate the scoring system into their
analyses. The expected data inputs are the \texttt{DataFrame} with the timestamp
and the anomaly score, and a \texttt{numpy.array} with the timestamp of the
start of the events of interest.

\section{Case of study}
\label{sec:results}

In this section, we include the details of the analysis carried out with a real
case of study to illustrate the appropriateness of the proposed measures and the
comparison between the different modules. This analysis includes a comparison
with the evaluation method for the most similar scenario found in the literature.

\subsection{Description of ArcelorMittal Sensor Data}
\label{sec:descr-arcel-shov}

The used data have been provided by ArcelorMittal. It comes from an asset that
requires permanent attention as failures occur with a high frequency. Depending
on the importance of the failure, the machine can be stopped for a quick repair
or need several days of reparation. The aim is to prevent these serious breakdowns
through the early detection of the machine problems.

The data includes the sensor time series and some other related information
(which may indicate some problem but do not imply an event of interest), \eg
failures logs, contextual information, etc. The data set consists of more than
38 million observations of 112 numeric attributes,
which involve information from operational and environmental
  contexts.

We have preprocessed the data, scaling it to the zero-one range to prevent an
artificial algorithm behaviour and discarding six features from a total of 112
due to their constant value.

\subsection{Algorithms involved in the experimentation}
\label{sec:algor-involv-exper}

The experimentation includes the results obtained by three unsupervised anomaly
detection algorithms. These algorithms are big data redesigns of some classic
anomaly detection algorithms and they are available in the
AnomalyDSD\footnote{https://spark-packages.org/package/ari-dasci/S-AnomalyDSD}
Spark Package. The use of big data algorithms is derived from the volume of the
used data set.

\begin{itemize}
  \item HBOS\_BD: Histogram-based Outlier Score (HBOS) anomaly detection
        algorithm~\cite{2012-Goldstein-Histogrambasedoutlierscore}. HBOS makes a
        histogram for every feature of the data to assign an anomaly score
        according to the number of instances present in each histogram bin. Two
        alternatives are proposed to process the numerical features: Static,
        with equal-width bins, and dynamic, where the values are sorted and
        divided in an equal number of instances bins.
  \item LODA\_BD: Lightweight Online Detector of Anomalies (LODA) is an
        ensemble-method based on the combination of random one-dimensional
        histograms~\cite{2016-Pevny-LodaLightweightonline}. The selection of
        random variables to make the histogram introduces a degree of
        variability, a desired feature in ensemble-based methods.
  \item XGBOD\_BD: Extreme Gradient Boosting Outlier Detection (XGBOD)
       ~\cite{2018-Zhao-XGBODimprovingsupervised} is an adaptation of XGBoost to
        a semi-supervised scenario. This algorithm uses unsupervised anomaly
        detection algorithms to obtain a representation for a supervised
        classifier.
\end{itemize}

In \autoref{tab:params} we include the default parameters for each algorithm
involved in the comparison. The best parameters for the algorithms has been
determined from an hyper-parameter optimization using the Optuna Optimization
Framework \cite{2019-Akiba-OptunaNextgenerationHyperparameter}. The used measure
for the optimisation has been the proposed ROC-AUC for a 6 hours window.

\begin{table}[htb]
		\centering
		\begin{tabular}{lp{90mm}}
			\toprule
			Algorithm & Parameters \\
			\midrule
			HBOS\_BD & n\_bins = 100, strategy =  ``static" \\

			LODA\_BD & n\_bins = 100, k = 100 \\


			XGBOD\_BD & detector = ``LODA\_BD", n\_TOS = 10, n\_selected\_TOS = 5, TOS\_strategy = ``acc", threshold = 0.1\\
			\bottomrule
		\end{tabular}
		\caption{Default parameter setting for the anomaly detectors}
		\label{tab:params}
\end{table}

These algorithms provide anomaly scores for each instance, so the filter
functions described in \autoref{sec:filt-cons} has not been used, although they
are included in the paper to provide the mechanisms to adapt the scoring system
to the researcher interests.

\subsection{Results and Analysis}
\label{sec:results-analysis}

In~\autoref{tab:auc} we include the AUC value for the different algorithms for
some values of intervals (1, 6 and 48 hours) using the different
aggregation functions. The best AUC result is highlighted in bold type. For this
evaluation method, HBOS variants are in general the most effective, pointing out
the general abnormality of time intervals previous to the alarms. The increasing
AUC value concerning the considered period is general to all aggregation and
weighting schemes. As described in \autoref{sec:methodology}, this is the
expected behaviour as there are fewer windows to be considered and those
previous to an incident is considered to be more abnormal. However, this
increase is not guaranteed for an algorithm with random performance, and such
AUCs are obtained here for certain algorithms and small windows. In this
particular study case, low AUCs should not be attributed to the random
performance, but to the relation between the anomalous behaviour of the machine
and the timestamp where the event of interest starts.

\begin{table}[htb!]
  \centering
  \begin{tabular}{cc|rrrrrr}
  \hline
 \toprule
  \multirow{2}{*}{Previous hours} & \multirow{2}{*}{Agg. function}
  & \multicolumn{2}{c}{HBOS} & \multirow{2}{*}{LODA} & \multirow{2}{*}{XGBOD} \\ 
			\cmidrule{3-4}
			& & Static   & Dynamic  &          &          \\ 
                                                               \midrule
    \multirow{3}{*}{1} & mean & 0.5623 & \textbf{0.5760} & 0.5592  & 0.4963 \\
                       & ccdf & 0.4206 & 0.4410 & 0.4252          & \textbf{0.4510} \\
                       & NAB & \textbf{0.5159} & 0.5155 & 0.5155    & 0.4314 \\
  \hline
    \multirow{3}{*}{6} & mean & 0.5692 & \textbf{0.5886} & 0.5584  & 0.4924 \\
                       & ccdf & 0.4619 & 0.4734 & 0.4827          & \textbf{0.5092} \\
                       & NAB & 0.6569 & \textbf{0.6571} & 0.6557  & 0.6258 \\
  \hline
   \multirow{3}{*}{48} & mean & \textbf{0.7577} & 0.7334 & 0.6781 & 0.5651 \\
                       & ccdf & 0.5792 & 0.5620 & \textbf{0.6418}& 0.6094 \\
                       & NAB & 0.8373 & 0.8322 & 0.8332 & \textbf{0.8537} \\
 \bottomrule
\end{tabular}

  \caption{AUC value for each algorithm and window length using different aggregations.}
  \label{tab:auc}
\end{table}


Concerning the Complementary CDF aggregation, the values show how a non-linear
aggregation benefits the XGBOD algorithm. Therefore, those algorithms are more
suitable for scenarios where single points may indicate an abnormal problem
instead of those where the anomalies come from a general degradation of the
series.


The values using the Numenta weighting mechanism are similar to the results
using the mean aggregation, as we are using the same aggregation function.
However, they are higher for wider windows, as the values further from the
annotated event have more relative weight and we have seen that the algorithms
do not perform well for narrow windows. It is important to note that XGBOD
obtains the best result for the 48 hours window length with this evaluation
scheme, so this algorithm also detects an abnormal general behaviour previous to
the event, although only at the start of the window, which is the performance
rewarded by the NAB scorer.

\begin{figure}[h]
    \begin{subfigure}{0.5\textwidth}
    \centering
    \includegraphics[width=\textwidth,clip]{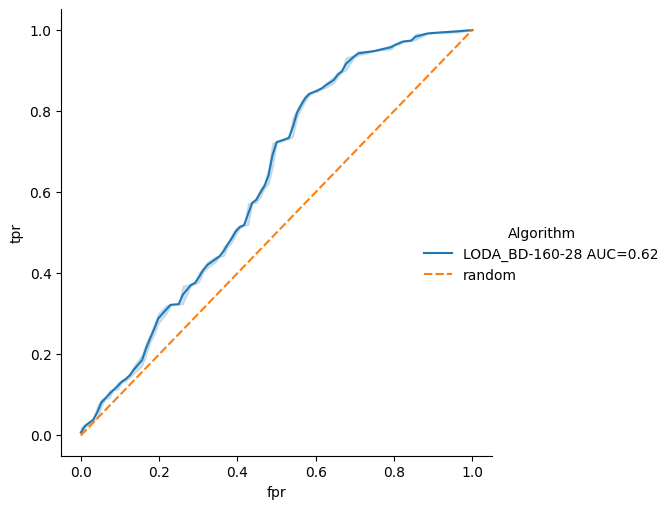}
    \caption{Example of ROC curve for 36 hour window}
    \label{fig:roc-curve}
  \end{subfigure}
\begin{subfigure}{0.5\textwidth}
  \centering
  \includegraphics[width=\textwidth,clip, trim=8cm 0cm 5cm 0cm]{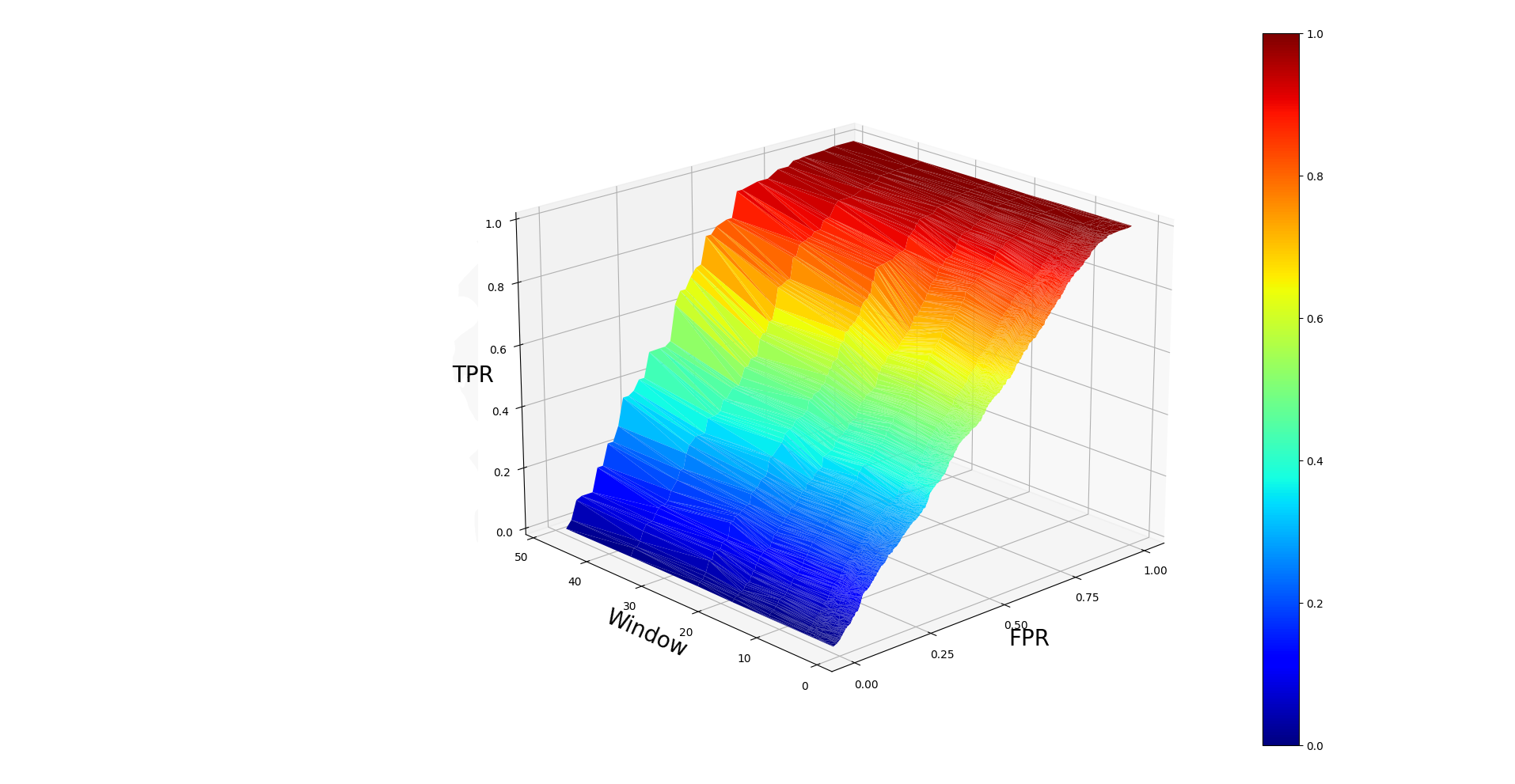}
  \caption{ROC Surface for LODA algorithm}
  \label{fig:roc-surface}
\end{subfigure}
  \caption{ROC evaluation}
  \label{fig:roc-evaluation}
  \end{figure}

  In \autoref{fig:roc-curve}, the ROC curve for the LODA algorithm with the mean
  aggregation for the 36 hours interval previous to an event is shown.
  The performance is only slightly better than a random prediction, although for
  this window length it could be useful in the detection of anomalies. In
  \autoref{fig:roc-surface} we show the ROC surface of the LODA algorithm for
  the windows up to 48 hours previous to an incident. The aggregation method
  used is the mean of the score values.

  In this image, the ROC surface shows that the performance is close to the random performance for windows shorter than 20 hours. We can observe that a surge in the performance when the window length is close to the upper limit, particularly in the longest window, where there is an increase in the TPR while the FPR is still low.

  We have performed statistical analysis with the results of the AUCs for the
  mean aggregation to compare the algorithms. This paper does not present an
  anomaly detection algorithm, but an evaluation framework, so the analyses are
  not centred on one of the algorithms. To prove that there are differences
  between results, we have used the Friedman test, as the results are not
  expected to come from Gaussian distribution due to the different window
  lengths~\cite{2020-Carrasco-Recenttrendsuse}. The obtained $p$-value is
  $1.41 \cdot 10^{-6}$, so we can reject the null hypothesis that represents the
  equivalence of the methods.

\begin{figure}[ht]
  \centering
  \includegraphics[width=\textwidth, clip, trim=4cm 3cm 4cm 2.2cm]{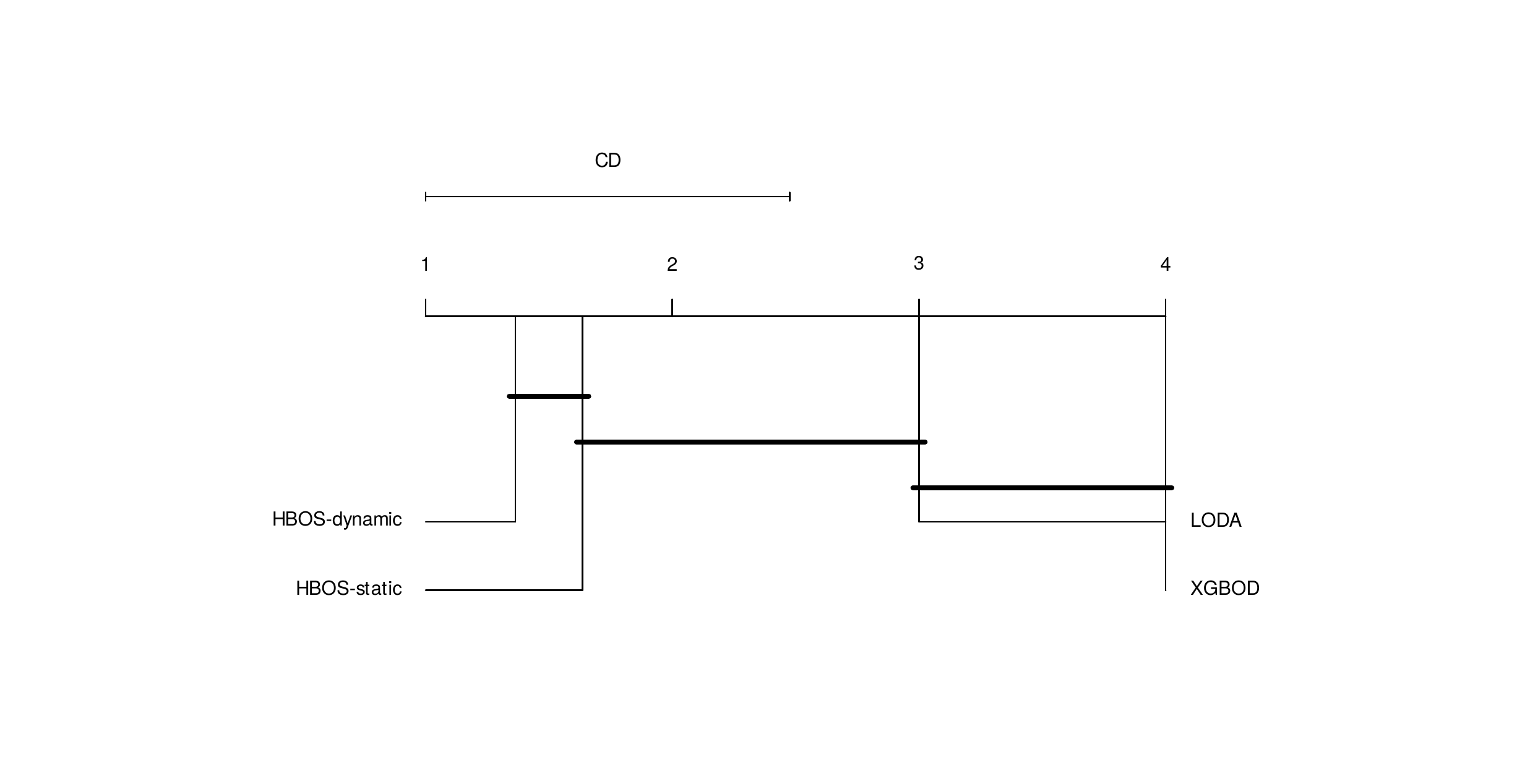}
  \caption{Critical Difference comparison plot}
  \label{fig:cd-plot}
\end{figure}

  The results of a Friedman test with the Holland Post-Hoc adjust, used to
  search for the differences claimed by the Friedman test, are shown graphically
  in \autoref{fig:cd-plot}. Here is shown that although HBOS variants are the
  algorithms with better results, we cannot discard the possibility that these
  differences are produced by chance. The main equivalence between the
  algorithms that can be discarded is between the dynamic variant of the HBOS
  algorithm and both LODA and XGBOD and between XGBOD and HBOS variants.


\subsection{Comparison to Range-Based Precision and Recall}
\label{sec:comp-rang-based}

In this section, we compare our method to the range-based scoring system
\cite{2019-Tatbul-PrecisionRecallTime}. The goal of the comparison is to show
that the concept of the preceding window ROC includes the information computed
by the range-based precision and recall, with the additional benefits of the ROC
curve for unsupervised scenarios. Some considerations should be made concerning
this comparison:

\begin{itemize}
  \item The range-based proposal is intended to evaluate the detection of
        anomalous intervals, so we have adapted the dataset, labelling as
        anomalous the instances previous to the events using different window
        lengths, similarly as in the proposed method.
  \item This method assumes that the outputs of the algorithm are anomaly
        labels, while our method works with the anomaly scores. Then, for the
        algorithms involved in the comparison, certain anomaly levels have been
        used to transform the score into dichotomic labels. For the sake of a
        more direct comparison, we have computed the precision, recall and $F_1$
        at equivalent thresholds of the proposed ROC curve. It is important to
        note that the thresholds in the pw-ROC evaluation refer to the
        aggregated scores, although there is not an univocal relation between
        instances.
  \item The comparison has been made using a subset due to the computational
        cost of the computation of the range-based measures.
  \item We have used the mean aggregation for the pw-ROC measures.
  \item The time-based bias described in the range-based proposal has a similar
        goal to the NAB weights. Here we have used a flat bias in both scoring
        systems to focus the comparison in the evaluation of the anomaly
        detection.
\end{itemize}

For each algorithm, let $q_{0.05}$ and $q_{0.95}$ be the 0.05 and 0.95 quantile
of the scores, respectively. Then, the labels has been assigned using a linear
mapping for the thresholds:
\begin{equation} \mathrm{Label}(x, \alpha) = \left\lbrace \begin{array}{ll} 1  & \mbox{if } x \geq  q_{0.05}+ \alpha(q_{0.95} - q_{0.05})\\ 0 & \mbox{otherwise} \end{array} \right., \end{equation}
where $\alpha$ takes the values of $0.2, 0.5$ and $0.8$. We have selected
the 0.05 and 0.95 quantiles to filter the most extreme score values. Then, we
have used three threshold to mimic a scenario where only an estimation of the
score distribution is available.

\begin{figure}[h]
  \centering
  \makebox[\textwidth][c]{\includegraphics[width=1.4\textwidth]{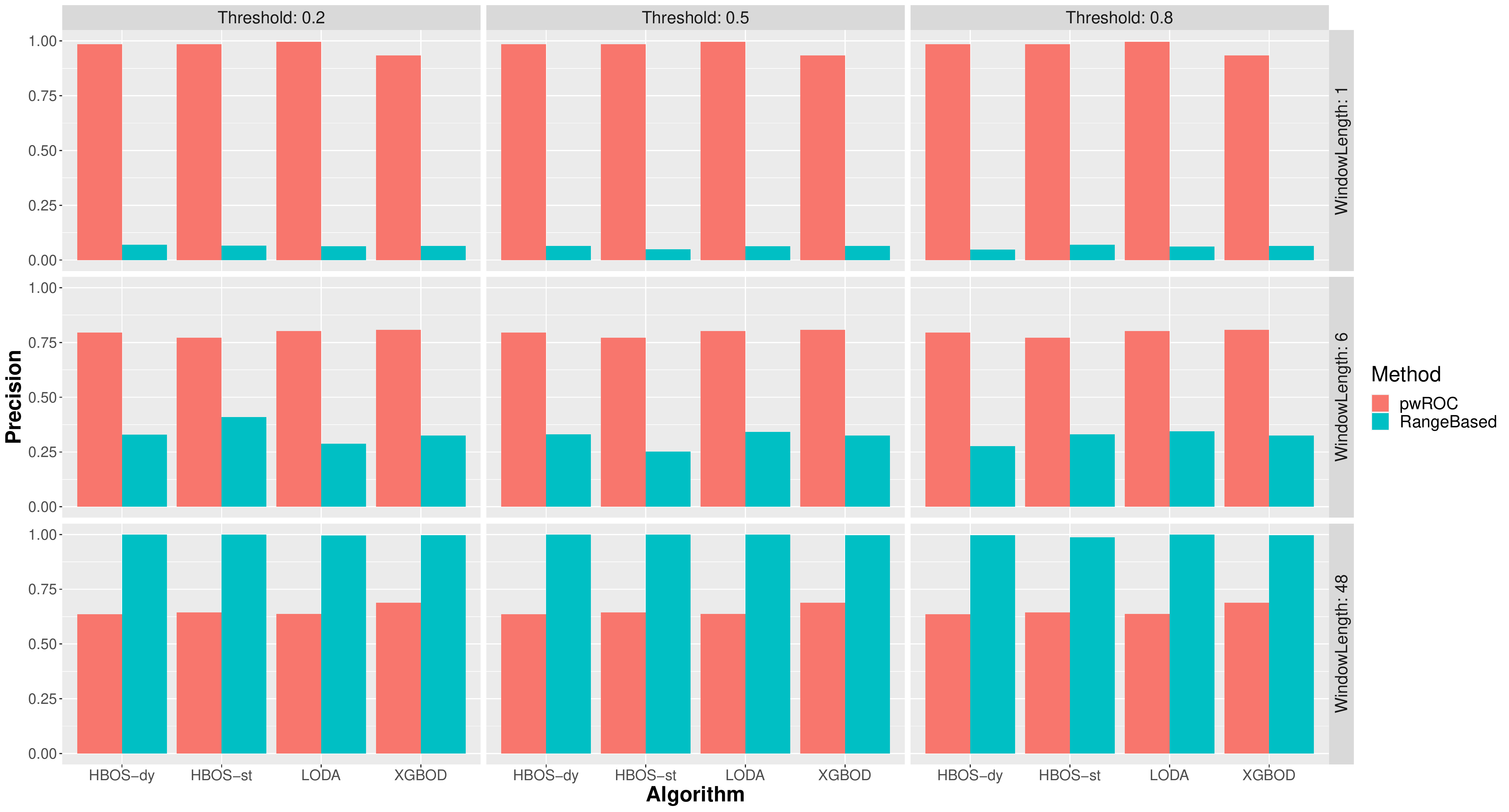}}
  \caption{Preceding Window vs Range-Based Precision}
  \label{fig:pw-rb-precision}
\end{figure}

\begin{figure}[h]
  \centering
  \makebox[\textwidth][c]{\includegraphics[width=1.4\textwidth]{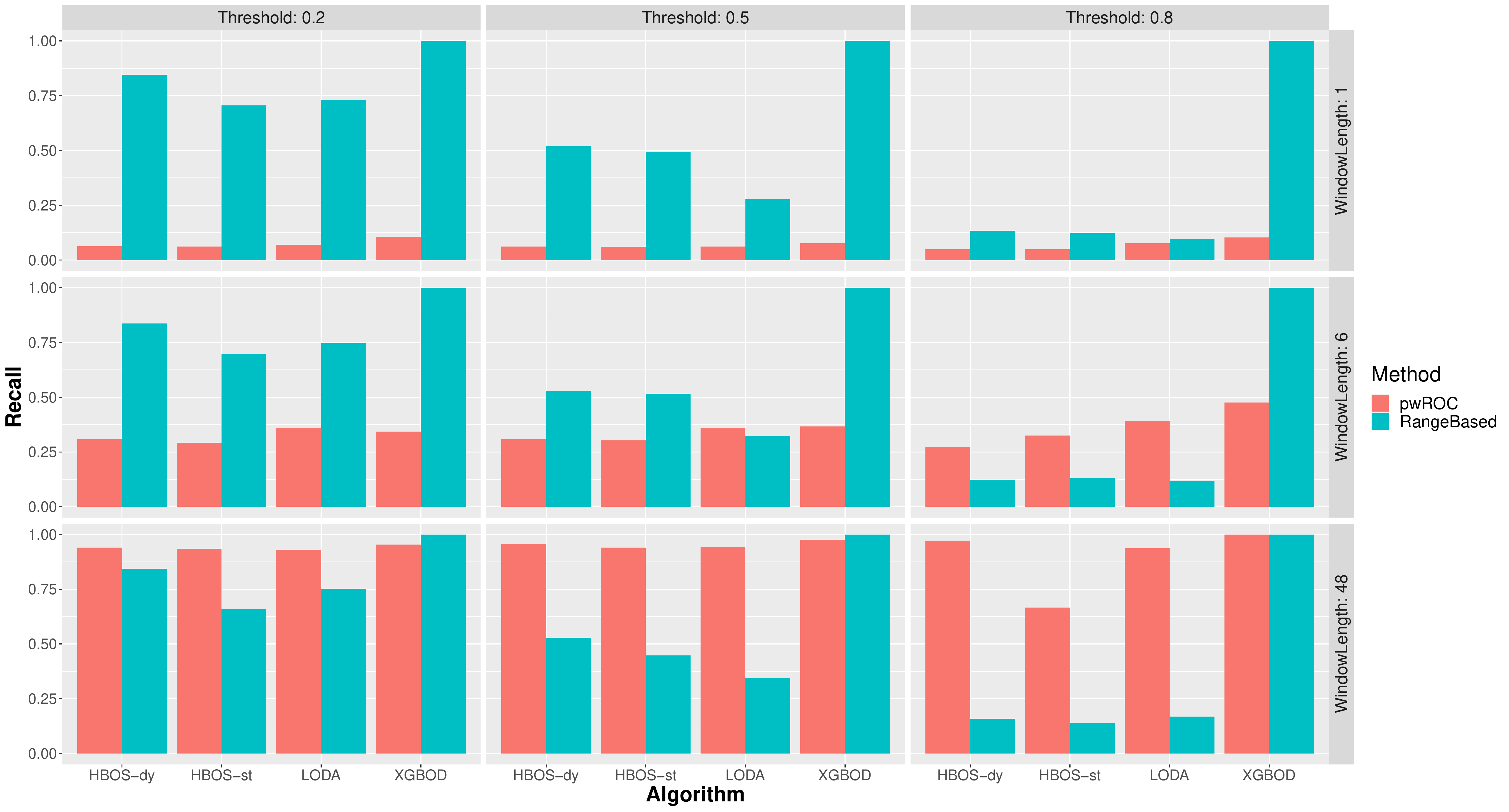}}
  \caption{Preceding Window vs Range-Based Recall}
  \label{fig:pw-rb-recall}
\end{figure}

\begin{figure}[h]
  \centering
  \makebox[\textwidth][c]{\includegraphics[width=1.4\textwidth]{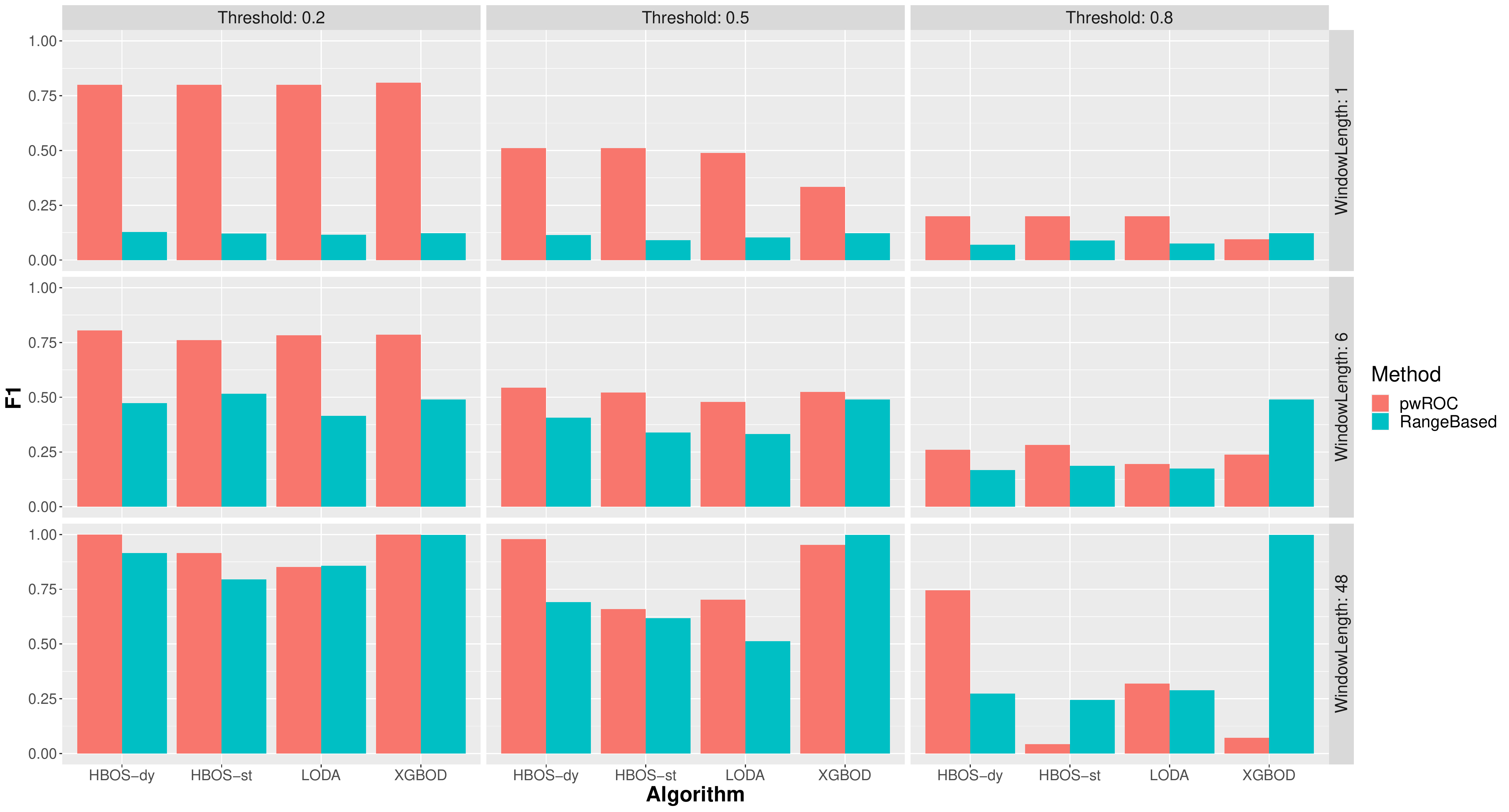}}
  \caption{Preceding Window vs Range-Based F1}
  \label{fig:pw-rb-f1}
\end{figure}

In~\autoref{tab:tatbul} we show the range-based $F_1$ score, using different
window lengths and thresholds. With the exception of the combination of the two
shortest windows and the lowest threshold, where the dynamic and static version
of HBOS get the best result for the 1 and 6 hours windows, the best $F_1$ score
is obtained in every other scenario by the XGBOD algorithm. The big differences
between the scores with different anomaly thresholds imply that this scoring
system rewards very positively the detection of anomalous instances without
penalising the false positives, which are very likely with a low threshold. The
differences with respect the pw-ROC evaluation, where XGBOD gets the worst
results, can be explained by the use of a threshold, that is determinant to the
detection of the anomalies. This circumstance is made clear by the following
\Crefrange{fig:pw-rb-precision}{fig:pw-rb-f1}, which show the comparison for the
precision, recall and $F_1$ measures between the range-based and the preceding
window for each threshold and window length for both scoring methods
(represented in colors).



\begin{table}
  \begin{tabular}{cc|rrrrrr}
  \hline
 \toprule
  \multirow{2}{*}{Previous hours} & \multirow{2}{*}{Threshold}
  & \multicolumn{2}{c}{HBOS} & \multirow{2}{*}{LODA} & \multirow{2}{*}{XGBOD} \\ 
			\cmidrule{3-4}
			& & Static   & Dynamic  &          &          \\ 
                                                               \midrule
    \multirow{3}{*}{1} & 0.2 & 0.121 & \textbf{0.129} & 0.116 & 0.122 \\
                       & 0.5 & 0.090 & 0.114 & 0.103 & \textbf{0.122} \\
                       & 0.8 & 0.089 & 0.071 & 0.075 & \textbf{0.122} \\
  \hline
    \multirow{3}{*}{6} & 0.2 & \textbf{0.516} & 0.473 & 0.415 & 0.490 \\
                       & 0.5 & 0.339 & 0.407 & 0.332 & \textbf{0.490} \\
                       & 0.8 & 0.188 & 0.168 & 0.175 & \textbf{0.490} \\
  \hline
   \multirow{3}{*}{48} & 0.2 & 0.795 & 0.915 & 0.857 & \textbf{0.998} \\
                       & 0.5 & 0.618 & 0.691 & 0.512 & \textbf{0.998} \\
                       & 0.8 & 0.245 & 0.274 & 0.289 & \textbf{0.998} \\
   \bottomrule
\end{tabular}

  \caption{Range-based $F_{1}$}
  \label{tab:tatbul}
\end{table}

\autoref{fig:pw-rb-precision} shows the precision of the algorithms for the
threshold and window lengths. The results for both scoring methods are more
similar for the precision measure. For the 48 window length, all the predicted
anomalous intervals fall into anomalous ranges, meaning a 1 range-based
precision. However, this is not the case for the pw-ROC computation of the
precision, as it may be that the mean scores of some positive windows are less
than the mean of instances outside the positive windows. Then, the situation
here could be that positive ranges and windows contain anomalous instances,
although the majority of scores are much lower.

The mentioned hypothesis is supported by the comparison between the range-based
recall and the recall within the previous window shown in \autoref{fig:pw-rb-recall}.
It is important to note that the range-based recall of the
  XGBOD algorithm is 1 in all the scenarios, which means that all the positive
  instances are labelled as anomalies for all the considered thresholds.
  Therefore, the determination of the anomaly threshold is for the evaluation of
  this algorithm, while the pw-ROC scoring system provides an evaluation more
  similar to the other algorithms ones. Then, for this approach, we should know
  the distribution of the scores and adjust the threshold accordingly, as the
  precision is very low for the lower window lengths. These results reinforce
  the hypothesis that ROC based scoring systems are fairer, as they show the
  effectiveness for all the possible thresholds. Another aspect that polarises
  the output of the ranged based evaluation for algorithms that have a narrow
  range of scores is the fact that the presence or absence of instances labelled
  as anomalies affect the evaluation of the whole interval, while the pw-ROC
  aggregates the information of the instances within the periods, a process that
  avoids the dichotomisation of the output.

The combination of the previous scores is shown in \autoref{fig:pw-rb-f1}, which
depicts the comparison between the range-based $F_1$ score for each threshold
and window length. The deepest difference here relies on the XGBOD algorithm,
which performs well in terms of the pw-ROC for the lower thresholds, but obtains
a near 1 range-based $F_1$ score for every threshold for the 48 hours window length.

As mentioned before, this is a case where the good definition of the threshold
is crucial, while the pw-ROC computation of the score is not affected by it, not
only by being unnecessary to the computation of the curve but through the window
aggregation.

\paragraph{Cost analysis}

The alleged computational complexity of the range-based measures is
$O(N_{r} \times N_{p})$, where $N_r$ is the number of real anomalous intervals and
$N_p$ represents the number of predicted anomalous intervals. However, this cost
omit the computation of the size of the intersection of the predicted and real
anomalous intervals and the biased weight of the instances within, which could
lead to a $O(n^2)$ computational cost, where $n$ is the number of instances. The
complexity of the proposed method is $O(n)$, which could be a greater number
than $N_r, N_p$, although our proposal is much more efficient in practice.

\autoref{tab:computation-time} shows the computational time of
  each evaluation method for each threshold and window length. The results
  included in this table are the mean computational time between the
  computational times for all the algorithms and it is measured in seconds. The
  computational cost of the range-based measure is much greater than the cost of
  the proposed method, especially when a larger window is considered. The
  increase in the computational cost needed by the range-based method for longer
  windows is derived from the need of computing the intersection between longer
  anomalous intervals. A similar circumstance happens with the 0.5 threshold,
  which implies a greater number of changes between normal and anomalous
  predicted intervals. On the contrary, the cost of the proposed method is only
  affected by the number of instances, as longer windows mean fewer aggregations
  of more instances.

\begin{table}[h]
  \centering
\begin{tabular}{ll|rr}
 \toprule
WindowLength        & Threshold & \multicolumn{1}{c}{pwROC} & \multicolumn{1}{c}{RangedBased} \\
  \midrule
\multirow{3}{*}{1}  & 0.2       & 0.14                      & 146.12                          \\
                    & 0.5       & 0.11                      & 214.73                          \\
                    & 0.8       & 0.09                      & 101.24                          \\ \hline
\multirow{3}{*}{6}  & 0.2       & 0.46                      & 233.41                          \\
                    & 0.5       & 0.41                      & 360.14                          \\
                    & 0.8       & 0.29                      & 167.15                          \\ \hline
\multirow{3}{*}{48} & 0.2       & 0.94                      & 17533.13                        \\
                    & 0.5       & 0.88                      & 29105.63                        \\
                    & 0.8       & 0.38                      & 12166.45 \\
 \bottomrule
\end{tabular}
\caption{Comparison of computational cost (sec) of evaluation methods}\label{tab:computation-time}
\end{table}

\section{Concluding remarks}
\label{sec:conclusion}

This work proposes an evaluation framework for anomaly detection for time series
scenarios. This framework allows the use of AUC, an extended anomaly detection
measure, for event detection regarding the uncertainty about the annotated
timestamp of the event. The description of the components applied to the
original time series data allows the researcher to propose new aggregation
functions designed for their particular case within the proposed framework,
which can lead to more appropriate conclusions. We also have adapted into the
framework the well-known Numenta Anomaly Benchmark scoring system to reward the
early detection of the anomalies although our definition avoids the
proliferation of parameters that can obscure the meaning of the measure.

The experimentation with three different distributed algorithms for a real-world
case study with the mentioned characteristics of anomaly annotation, and the
comparison with a range-based scoring system, have validated the robustness of
the proposal of the used metrics and methods through the description of a ROC
based score instead of depending on an anomaly threshold, which could bring a
more dichotomous situation. Therefore, the presented evaluation method has made
viable the study of a problem of anomaly detection for time series, with both
classic and Big Data implementations of the scoring method, which otherwise
would have to use inappropriate or inefficient measures that could lead to wrong
conclusions about the quality of the algorithms.

\section*{Acknowledgements}
\label{sec:acknowledgements}

This work has been partially supported by the Ministry of Science and Technology
under project TIN2017-89517-P, the Contract UGR-AM OTRI-4260 and the Andalusian
Excellence project P18-FR-4961. J. Carrasco was supported by the Spanish
Ministry of Science under the FPU Programme 998758-2016. D. García-Gil holds a
contract co-financed by the European Social Fund and the Administration of the
Junta de Andalucía, reference DOC\_01137.

\bibliography{bibliography.bib}

\end{document}